\documentclass[10pt,twocolumn,letterpaper]{article}
\usepackage[accsupp]{axessibility}
\usepackage{iccv}
\usepackage{times}
\usepackage{epsfig}
\usepackage{graphicx}
\usepackage{amsmath}
\usepackage{amssymb}
\usepackage{float}
\usepackage{booktabs}
\usepackage{multirow}
\usepackage{rotating}
\usepackage{bbding}
\usepackage{enumitem}
\usepackage{bm}
\usepackage[ruled,vlined]{algorithm2e}
\usepackage{times}
\usepackage{epsfig}
\usepackage{graphicx}
\usepackage{amsmath}
\usepackage{amssymb}
\usepackage{array}
\usepackage{subcaption}

\newcolumntype{x}[1]{>{\centering\arraybackslash}p{#1pt}}
\newcommand{\tablestyle}[2]{\setlength{\tabcolsep}{#1}\renewcommand{\arraystretch}{#2}\centering}

\usepackage[dvipsnames]{xcolor}
\usepackage{colortbl} 
\definecolor{mypink}{cmyk}{0, 0.7808, 0.4429, 0.1412}
\definecolor{mygreen}{rgb}{0.1725, 0.7059, 0.3921}
\definecolor{mypurple}{rgb}{0.396, 0.1601, 0.5764}
\definecolor{myblue}{rgb}{0.0, 0.72, 0.92}
\definecolor{mybrown}{rgb}{0.7176, 0.5294, 0.1098}
\definecolor{mygray}{gray}{0.6}
\definecolor{mygray-bg}{gray}{0.9}
\definecolor{mytomato}{rgb}{0.941, 0.4276, 0.3138}
\usepackage{graphicx}

\makeatletter
\newcommand*\bigcdot{\mathpalette\bigcdot@{.5}}
\newcommand*\bigcdot@[2]{\mathbin{\vcenter{\hbox{\scalebox{#2}{$\m@th#1\bullet$}}}}}
\makeatother

\newcommand\blfootnote[1]{%
  \begingroup
  \renewcommand\thefootnote{}\footnote{#1}%
  \addtocounter{footnote}{-1}%
  \endgroup
}

\usepackage[pagebackref=true,breaklinks=true,letterpaper=true,colorlinks,bookmarks=false]{hyperref}
\iccvfinalcopy 

\def\iccvPaperID{****} 

\ificcvfinal\fi

\begin{document}

\title{Compositional Feature Augmentation for Unbiased Scene Graph Generation}

\author{Lin Li$^{1,2}$,
    Guikun Chen$^1$,
    Jun Xiao$^1$,
    Yi Yang$^1$,
    Chunping Wang$^3$,
    Long Chen$^{2\dagger}$ \\
    \small{ $^1$Zhejiang University \;\;
            $^2$The Hong Kong University of Science and Technology \;\;
            $^3$FinVolution} \\
  \small{\texttt{\{mukti, guikun.chen, junx, yangyics\}@zju.edu.cn, wangchunping02@xinye.com, longchen@ust.hk}} \\
  \small{\href{https://github.com/HKUST-LongGroup/CFA}{https://github.com/HKUST-LongGroup/CFA}}
}

\maketitle
\ificcvfinal\thispagestyle{empty}\fi

\begin{abstract}
Scene Graph Generation (SGG) aims to detect all the visual relation triplets $<$\texttt{sub}, \texttt{pred}, \texttt{obj}$>$ in a given image. With the emergence of various advanced techniques for better utilizing both the intrinsic and extrinsic information in each relation triplet, SGG has achieved great progress over the recent years. However, due to the ubiquitous long-tailed predicate distributions, today's SGG models are still easily biased to the head predicates. Currently, the most prevalent debiasing solutions for SGG are re-balancing methods, \eg, changing the distributions of original training samples. In this paper, we argue that all existing re-balancing strategies fail to increase the diversity of the relation triplet features of each predicate, which is critical for robust SGG. To this end, we propose a novel Compositional Feature Augmentation (\textbf{CFA}) strategy, which is the first unbiased SGG work to mitigate the bias issue from the perspective of increasing the diversity of triplet features. Specifically, we first decompose each relation triplet feature into two components: intrinsic feature and extrinsic feature, which correspond to the intrinsic characteristics and extrinsic contexts of a relation triplet, respectively. Then, we design two different feature augmentation modules to enrich the feature diversity of original relation triplets by replacing or mixing up either their intrinsic or extrinsic features from other samples. Due to its model-agnostic nature, CFA can be seamlessly incorporated into various SGG frameworks. Extensive ablations have shown that CFA achieves a new state-of-the-art performance on the trade-off between different metrics.
\end{abstract}
\blfootnote{$^\dagger$ Corresponding author. Work was done when Lin Li visited HKUST.}
\section{Introduction}

As one of the fundamental comprehensive visual scene understanding tasks, Scene Graph Generation (\textbf{SGG}) has attracted unprecedented interest from our community and has made great progress in recent years~\cite{lu2016visual, zhang2017visual, zellers2018neural, chen2019counterfactual, chen2019knowledge, tang2020unbiased,li2023label,li2021bipartite, li2022devil,yu2023visually,li2023zero,ohashi2022unbiased}. Specifically, SGG aims to transform an image into a visually-grounded graph representation (\ie, scene graph) where each node represents an object instance with a bounding box and each directed edge represents the corresponding predicate between the two objects. Thus, each scene graph can also be formulated as a set of visual relation triplets (\ie, $<$\texttt{sub}, \texttt{pred}, \texttt{obj}$>$). Since such structural representations can provide strong explainable potentials, SGG has been widely-used in various downstream tasks, such as visual question answering, image retrieval, and captioning.

\begin{figure}[!t]
  \centering
  \includegraphics[width=\linewidth]{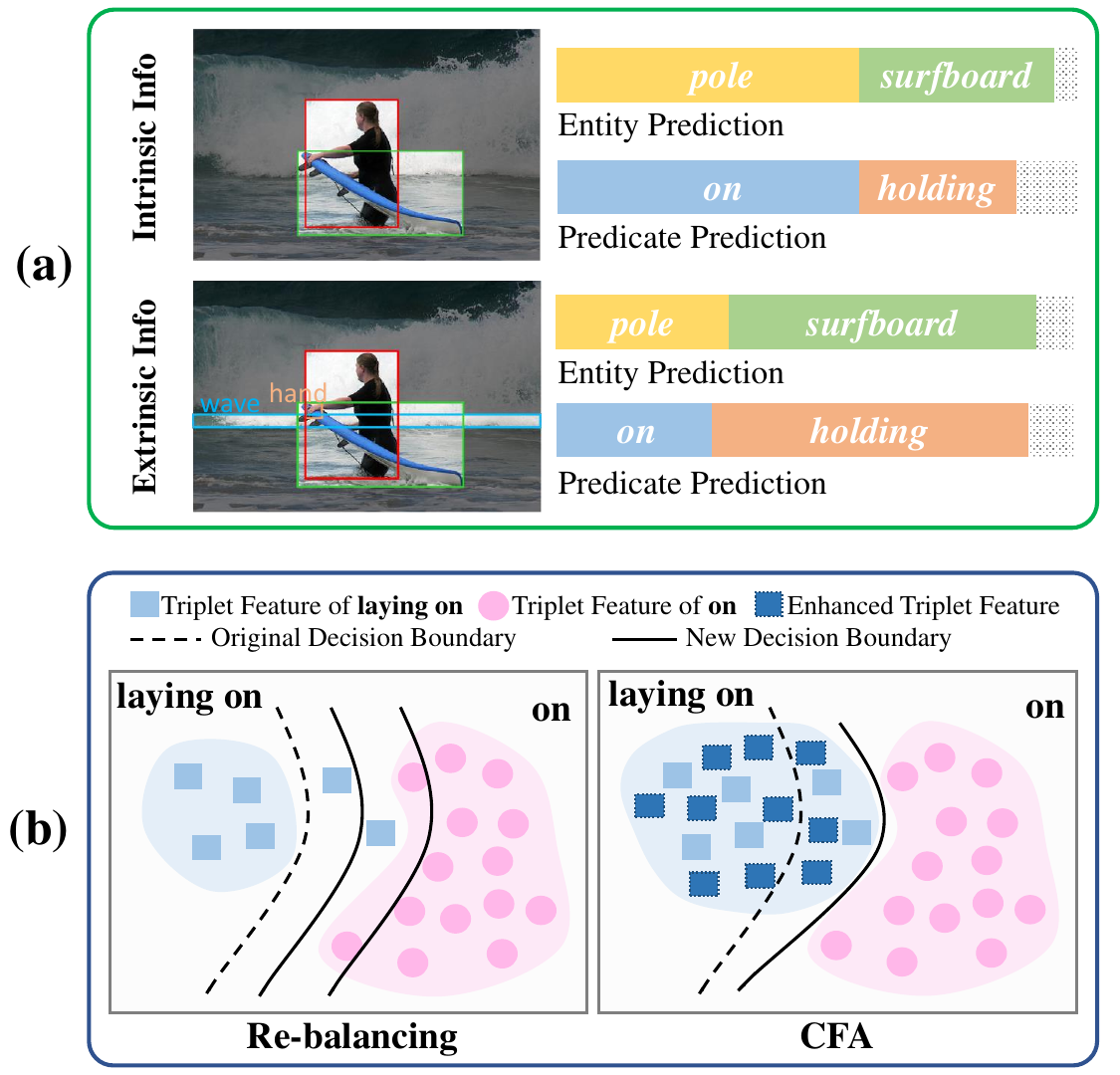}
   
  \caption{\textbf{(a)} The intrinsic and extrinsic information for SGG. The entity prediction is for the \textcolor{Green}{green} box, and the predicate prediction is for the relation between the \textcolor{red}{red} and \textcolor{Green}{green} boxes. \textbf{(b)} Illustration of the diversity of feature space and decision boundary between \texttt{on} and \texttt{laying on} before and after using re-balancing and CFA. Each sample denotes the corresponding visual triplet features.}
  \label{fig:motivation}
\end{figure}

In general, due to the extremely diverse visual appearance of different visual relation triplets, recent SGG methods all consider both \textbf{\emph{intrinsic}} and \textbf{\emph{extrinsic}} information for entity and predicate classification~\cite{zellers2018neural,tang2019learning,lu2021context}. By ``intrinsic information", we mean these intrinsic characteristics of the subjects and objects, such as their visual, semantic, and spatial features. For example in Figure~\ref{fig:motivation}(a), with the help of these intrinsic features, we can easily infer all possible entity categories (\eg, \texttt{pole} or \texttt{surfboard}) and predicate categories (\eg, \texttt{on} or \texttt{holding}) of each triplet. However, sometimes it is still hard to confirm the exact correct predictions with only intrinsic information, especially for tiny objects. Thus, it is also essential to consider other ``extrinsic information" in the same image, such as the context features from neighbor objects. As shown in Figure~\ref{fig:motivation}(a), after encoding the features of surrounding objects (\eg, \texttt{wave} and \texttt{hand}), we can easily infer that the categories of the entity and predicate should be \texttt{surfboard} and \texttt{holding}.

Although numerous advanced techniques have been proposed to effectively leverage both intrinsic and extrinsic information, today's SGG methods still fail to predict some informative predicates due to the ubiquitous long-tailed predicate distribution in prevalent SGG datasets~\cite{krishna2017visual}. Such a distribution is characterized by few categories with vast samples (head\footnote{We directly use “tail”, “body”, and “head” categories to represent the predicate categories in the tail, body, and head parts of the number distributions of different predicates in SGG datasets, respectively.}) and many categories with rare samples (tail). Since the discrepancy of feature diversity and sample size among different categories, the learned decision boundary becomes improper (\cf.~Figure~\ref{fig:motivation}(b)), \ie, their predictions are biased towards the head predicates (\eg, \texttt{on}) and they are error-prone for the tail ones (\eg, \texttt{laying on}).

To overcome the bias issue, the most prevalent unbiased SGG solutions are re-balancing strategies, \eg, sample re-sampling~\cite{li2021bipartite,zhou2022peer} and loss re-weighting~\cite{yan2020pcpl,chen2022resistance,li2022ppdl,lyu2022fine,kang2023skew}. They alleviate the negative impact of long-tailed distribution by increasing samples or loss weights of tail classes. Then, the decision boundaries are adjusted to reduce the bias introduced by imbalanced distributions. However, we argue that all the existing re-balancing strategies fail to increase the diversity of relation triplet features\footnote{We use the ``relation triplet feature" to represent the combination of both \textbf{intrinsic} and \textbf{extrinsic} features of each visual relation triplet. \label{ft:triplet_feat}} of each predicate, \ie, they only change the frequencies or contributions of existing relation triplet features (\cf.~Figure~\ref{fig:motivation}(b)). Since these tail categories are under-represented, it is still hard to infer the complete data distribution, \ie, making it challenging to find the optimal direction to adjust the decision boundaries~\cite{chu2020feature,vigneswaran2021feature}. For example in Figure~\ref{fig:motivation}(b), the feature space of \texttt{laying on} is so sparse that the decision boundary can be adjusted within a large range. The performance of this naive adjustment without ``complete" distribution is always sensitive to hyperparameters, \ie, excessively increasing the sample number or loss weight of tail predicates may cause some head predicate samples to be incorrectly predicted as tail classes (right solid line), and vice versa (left solid line).

\begin{figure}[!t]
  \centering
  \includegraphics[width=\linewidth]{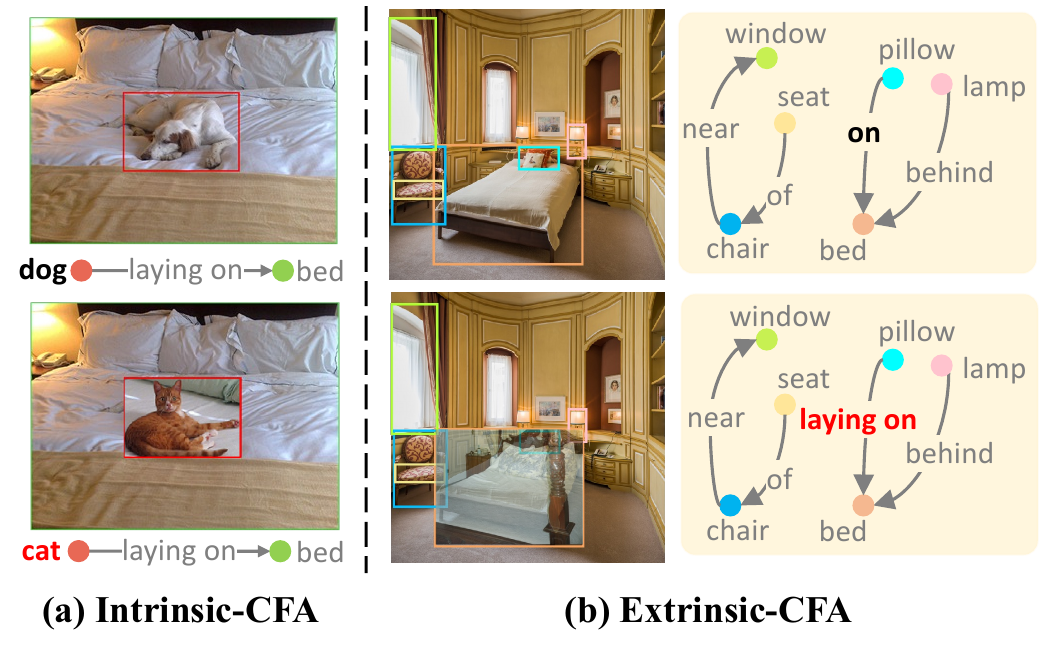}
  \vspace{-1.5em}
  \caption{\textbf{(a) Intrinsic-CFA}: Replacing the entity feature of tail predicate triplet \texttt{dog}-\texttt{laying on}-\texttt{bed} from \texttt{dog} to \texttt{cat} to enhance the intrinsic feature. \textbf{(b) Extrnisc-CFA}: Mixing up the feature of tail predicate triplet \texttt{pillow}-\texttt{laying on}-\texttt{bed} into the context of \texttt{pillow}-\texttt{on}-\texttt{bed} to enhance the extrinsic feature.}
  \label{fig:component}
\end{figure}

In this paper, we propose a novel Compositional Feature Augmentation (CFA) strategy for unbiased SGG, which tries to solve the bias issue by enhancing the diversity of relation triplet features, especially for tail predicates. Specifically, CFA consists of two components: an intrinsic feature augmentation (\emph{intrinsic-CFA}) and an extrinsic feature augmentation (\emph{extrinsic-CFA}), which enhance the intrinsic and extrinsic features\footref{ft:triplet_feat}, respectively. As shown in Figure~\ref{fig:motivation}(b), by increasing the feature diversity of the tail predicates (\eg, \texttt{laying on}), SGG models can easily learn the proper decision boundaries (vs. re-balancing strategies).

\textbf{For intrinsic-CFA}, we replace the entity features (\eg, subject or object) of a tail predicate triplet with other ``suitable" entity features. To determine the suitable entity categories for the augmentation, we propose a new hierarchical clustering method to find the correlations between different entity categories, and then we regard the entity features from the same cluster are suitable. Specifically, we calculate the category correlation by pattern, context, and semantic similarities. For example in Figure~\ref{fig:component}(a), the entity categories \texttt{cat} and \texttt{dog} are in the same cluster, \ie, we can augment the intrinsic feature by replacing the \texttt{dog} entity feature with a \texttt{cat} entity feature. \textbf{For extrinsic-CFA}, we take advantage of the context or interactions of other triplets (\ie, context triplets) and enhance the features of tail predicate triplets by these context triplets. Specifically, given a context triplet randomly selected from an image, we first select a reasonable tail predicate triplet as the target by limiting the categories and relative position of two objects. Then, to minimize the impact on the prediction of other triplets in the original image and make use of the extrinsic features of the context triplet, we use a \emph{mixup operation} to fuse the features of targeted tail predicate triplet into the context triplet. For example in Figure~\ref{fig:component}(b), triplet \texttt{pillow}-\texttt{laying on}-\texttt{bed} is mixed up into the image of \texttt{pillow}-\texttt{on}-\texttt{bed}.

We evaluate CFA on two most prevalent and challenging SGG datasets: Visual Genome (\textbf{VG})~\cite{krishna2017visual} and \textbf{GQA}~\cite{hudson2019gqa}. Since CFA is a model-agnostic debiasing strategy, it can be seamlessly incorporated into various SGG architectures\footnote{Following the mainstream and concurrent unbiased SGG works, we also only focus on two-stage frameworks. As for one-stage models, a similar idea can be applied at the image-level, and we leave it for future works. \label{ft:two_stage}} and consistently improve their performance. Unsurprisingly, CFA can achieve a new state-of-the-art performance on the trade-off between different metrics. Extensive ablations and results on multiple SGG tasks and backbones have shown the generalization ability and effectiveness of CFA.

In summary, we make three contributions in this paper:
  
\begin{enumerate}[leftmargin=4mm]
    \itemsep-0.4em
    \item We reveal the issue of existing re-balancing methods, \ie, the lack of triplet feature diversity of tail categories. To this end, we are the first to tackle unbiased SGG from the perspective of increasing the diversity of triplet features.
    \item We propose the model-agnostic CFA for unbiased SGG, which is an efficient and novel compositional learning framework that spans the feature space of the tail categories by two independent plug-and-play modules.
    \item Extensive results show the effectiveness of CFA, \ie, it achieves a new SOTA performance on  SGG benchmarks.
\end{enumerate}

\section{Related Work}

\noindent\textbf{Unbiased Scene Graph Generation.} Biased predictions prevent further use of scene graphs in real-world applications. Recent unbiased SGG works can be roughly divided into three main categories: 1) \emph{Re-balancing}: It alleviates the negative impact of long-tailed predicate distribution by re-weighting or re-sampling~\cite{yan2020pcpl,li2022ppdl,lyu2022fine,li2021bipartite, desai2021learning}. 2) \emph{Unbiased Inference}: It makes unbiased predictions based on biased models~\cite{tang2020unbiased,yu2021cogtree}. 3) \emph{Noisy Label Learning}: It reformulates SGG as a noisy label learning problem and corrects these noisy samples~\cite{li2022devil,li2022nicest}. In this work, we point out the drawbacks of existing re-balancing methods and study unbiased SGG from the new perspective of feature augmentation. 

\noindent\textbf{Feature Augmentation.} Data augmentation is a prevalent training trick to improve models' performance. Conventional data augmentation methods~\cite{devries2017improved,zhong2020random} usually synthesize new samples in a hand-crafted manner. Compared to these image-level data augmentation methods, feature augmentation is another efficient way to improve models' generalizability by directly synthesizing samples in the feature space~\cite{chu2020feature,li2021simple}. Compared to existing data augmentation applications, SGG is a sophisticated task that involves intrinsic and extrinsic features. In this work, we propose CFA to enrich the diversity of relation triplet features for debiasing.

\noindent\textbf{Compositional Learning (CL).}
CL has been successfully applied to various computer vision tasks. As for visual scene understanding, some Human Object Interaction detection works~\cite{hou2020visual,kato2018compositional,hou2021detecting} compose new interaction samples that significantly benefit both low-shot and zero-shot settings. To the best of our knowledge, only two unpublished SGG work~\cite{he2021semantic,knyazev2021generative} also uses CL. Compared to~\cite{he2021semantic}, we have several key differences: 1) They aim to generate new representations which are close to the original triplet. Instead, we try to increase the diversity of triplet features. 2) They only change the entities and ignore the extrinsic features. 3) Their augmentation strategies are mainly based on the spatial locations or IoU of the entities. The second work~\cite{knyazev2021generative} primarily addresses classification tasks. Our work differs from it in two aspects: 1) CFA increases the diversity of tail predicate features by leveraging both intrinsic and extrinsic information. 2) we aim to improve robustness and performance under long-tailed predicate distributions.

\begin{figure}[!t]
  \centering
  \includegraphics[width=\linewidth]{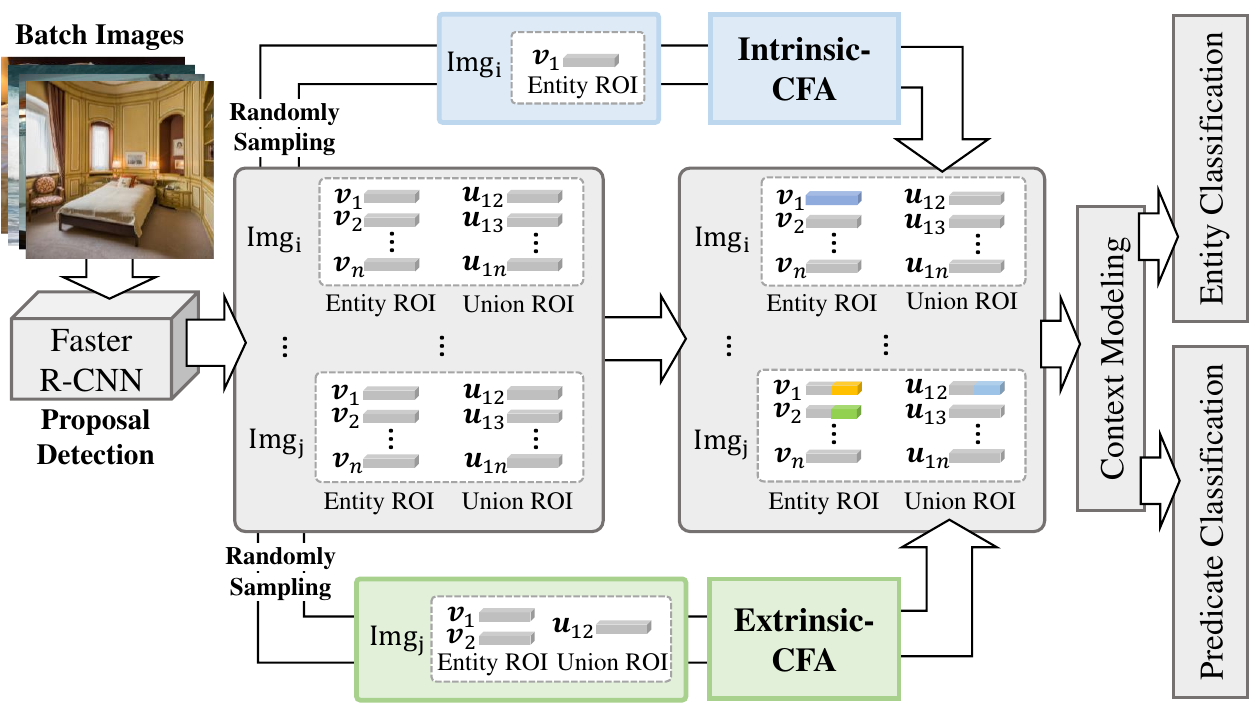}
  \vspace{-1em}
  \caption{The illustration of unbiased SGG framework with CFA.}
  \label{fig:framework}
\end{figure}

\begin{figure*}[!t]
  \centering
  \includegraphics[width=0.99\linewidth]{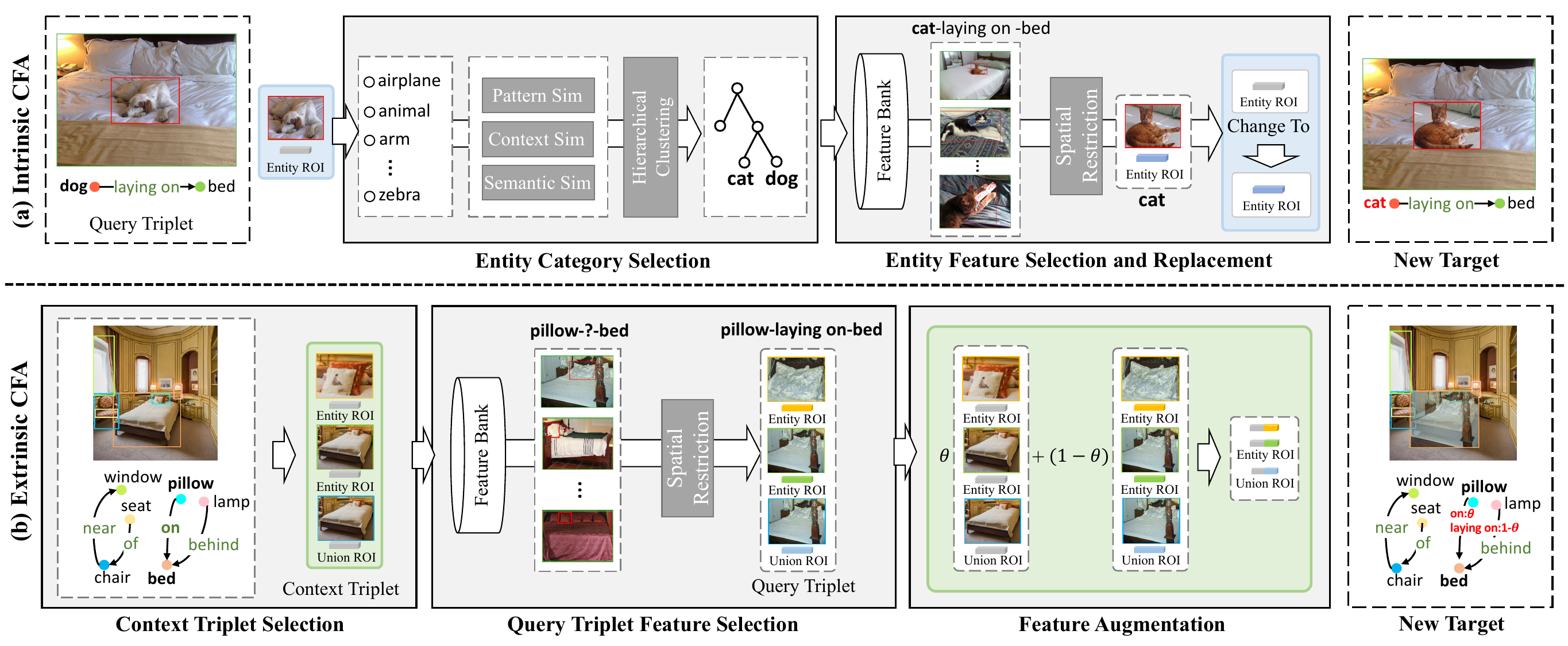}
   \vspace{-0.5em}
  \caption{The pipeline of Intrinsic CFA (a) and Extrinsic CFA (b). The blue and green boxes represent operations on the query triplet features in Intrinsic-CFA and context triplet features in Extrinsic-CFA, respectively.}
  \label{fig:cfa}
\end{figure*}

\section{Approach}

Given an image $I$, a scene graph is formally
represented as $\mathcal{G}$ = \{$\mathcal{N}$, $\mathcal{E}$\}, where $\mathcal{N}$ and $\mathcal{E}$ denote the set of all objects and their pairwise visual relations, respectively. Specifically, the $i$-th object in $\mathcal{N}$ consists of a bounding box (bbox) $\bm{b}_i \in \mathcal{B}$ and its entity category $o_i \in \mathcal{O}$. A relation $r_{ij} \in \mathcal{R}$ denotes the predicate category between $i$-th object and $j$-th object. $\bm{b}_{ij}$ denotes the union box of bbox $\bm{b}_i$ and bbox $\bm{b}_j$. $\mathcal{B}$, $\mathcal{O}$, and $\mathcal{R}$ represent the set of all entity bboxes, entity categories, and predicate categories, respectively.

In this section, we first revisit the two-stage SGG baselines in Sec.~\ref{sec:3.1}. Then, we detailedly introduce our CFA in Sec.~\ref{sec:cfa}, including intrinsic-/extrinsic-CFA (\cf.~Figure~\ref{fig:framework}). Finally, we introduce the training objectives in Sec.~\ref{sec:train}.

\subsection{Revisiting the Two-Stage SGG Baselines} \label{sec:3.1}

Since the mainstream SGG methods are two-stage models, we review the two-stage SGG framework\footref{ft:two_stage} here~\cite{zellers2018neural,tang2019learning}. A typical two-stage SGG model involves three steps: proposal generation, entity classification, and predicate classification. Thus, the SGG task $P(\mathcal{G}|I)$ is decomposed into:
\begin{equation}
P(\mathcal{G}|I)=P(\mathcal{B}|I)P(\mathcal{O}|\mathcal{B},I)P(R|\mathcal{O},\mathcal{B},I).
\end{equation}
\noindent\textbf{Proposal Generation $P(\mathcal{B}|I)$.} This step aims to generate all bbox proposals $\mathcal{B}$. Given an image $I$, they first utilize an off-the-shelf object detector (\eg, Faster R-CNN~\cite{ren2015faster}) to detect all the proposals $\mathcal{B}$ and their visual features $\{\bm{v}_i\}$.

\noindent\textbf{Entity Classification $P(\mathcal{O}|\mathcal{B},I)$.} This step mainly predicts
the entity category of each $\bm{b}_i$ $\in$ $\mathcal{B}$. Given a visual feature $\bm{v}_i$ and proposal $\bm{b}_i$, they use an object context encoder $\mathtt{Enc}_{obj}$ to extract the contextual entity representation $\bm{f}_i$: 
\begin{equation}
    \bm{f}_i = \mathtt{Enc}_{obj}(\bm{v}_i \oplus \bm{b}_i),
\end{equation}
where $\oplus$ denotes concatenation. Then, they use an object classifier $\mathtt{Cls}_{obj}$ to predict their entity categories:
\begin{equation}
    \widehat{o}_i = \mathtt{Cls}_{obj}(\bm{f}_i).
\end{equation}
\noindent\textbf{Predicate Classification $P(\mathcal{R}|\mathcal{O},\mathcal{B},I)$.} This step predicts the predicate categories of every two proposals in $\mathcal{B}$ along
with their entity categories. First, they use a relation context encoder $\mathtt{Enc}_{rel}$ to extract the refined entity feature $\bm{\tilde{f}}_i$:
\begin{equation}
    \bm{\tilde{f}}_i = \mathtt{Enc}_{rel}(\bm{v}_i \oplus \bm{f}_i \oplus \bm{w}_i),
\end{equation}
where $\bm{w}_i$ is the GloVe embedding~\cite{pennington2014glove} of predicted $\widehat{o}_i$. It is worth noting that both $\mathtt{Enc}_{cls}$ and $\mathtt{Enc}_{rel}$ often adopt a sequence model (\eg, Bi-LSTM~\cite{zellers2018neural}, Tree-LSTM~\cite{tang2019learning}, or Transformer~\cite{yu2021cogtree}) to better capture context. After relation feature encoding, they use a relation classifier $\mathtt{Cls}_{rel}$ to predict the relation $\widehat{r}_{ij}$ between any two subject-object pairs: 
\begin{equation}
    \widehat{r}_{ij} = \mathtt{Cls}_{rel}([\bm{\tilde{f}}_i \oplus \bm{\tilde{f}}_j] \circ \bm{u}_{ij}),
\end{equation}
where $\circ$ denotes element-wise product, and $\bm{u}_{ij}$ denotes the visual feature of the union box $\bm{b}_{ij}$.

\subsection{CFA: Compositional Feature Augmentation}
\label{sec:cfa}

In this paper, we treat a relation triplet feature\footref{ft:triplet_feat} in SGG as a combination of two components: intrinsic feature and extrinsic feature, which refer to the intrinsic information (\ie, the subject and object itself) and extrinsic information (\ie, the contextual objects and stuff), respectively. Correspondingly, our CFA consists of intrinsic-CFA and extrinsic-CFA to augment these two types of features. For ease of presentation, we call the ``targeted tail predicate triplet whose feature to be enhanced" as ``\textbf{query triplet}", and the ``relation triplet whose image is used to provide context" as ``\textbf{context triplet}". Besides, to facilitate the feature 
augmentation, we store all the visual features of all tail predicate triplets (\ie, visual features $\bm{v}$ of two entities and their union feature $\bm{u}$) in a \textbf{feature bank} before training. Besides, we adopt the repeat factor $\eta$ = $\max(1, \eta_r)$~\cite{gupta2019lvis,li2021bipartite} to sample images to provide enough context triplet and query triplet for augmentation, where $\eta_r$ = $\sqrt {\lambda/{f_r}}$, $f_r$ is the frequency of predicate category $r$ on the entire dataset, and $\lambda$ is hyperparameter. The unbiased SGG pipeline with CFA is shown in Figure~\ref{fig:framework}.

\subsubsection{Intrinsic-CFA}

Intrinsic-CFA enhances the feature of query triplet by replacing its visual features of entities (\ie, subject and object). It consists of two steps: \emph{entity category selection} and \emph{entity feature selection and replacement} (\cf.~Figure~\ref{fig:cfa}(a)). During training, we randomly select a query triplet from a batch of images, and randomly select one of its entity features as input. Then we put it into the Intrinsic-CFA module (\cf.~Figure~\ref{fig:framework}). Next, we detailedly introduce each step.

\noindent\textbf{Entity Category Selection.}
This step is used for determining the category to which the entity feature of query triplet is replaced. Firstly, we propose a novel \textbf{hierarchical clustering} strategy to mine potentially fungible entity categories. Then we randomly select an entity category from the same cluster for the next entity feature selection. For the query triplet \texttt{dog}-\texttt{laying on}-\texttt{bed} in Figure~\ref{fig:cfa}, the category \texttt{cat} is selected from the same cluster as \texttt{dog}. The categories in the same cluster are common in the behavior patterns (\eg, they can be ridden), contexts (\eg, they can appear in a scene with a street) as well as semantics. Accordingly, three kinds of similarity are used to measure the common characters of two entity categories: pattern similarity, context similarity, and semantic similarity.

\begin{figure}[!t]
  \centering
  \includegraphics[width=0.95\linewidth]{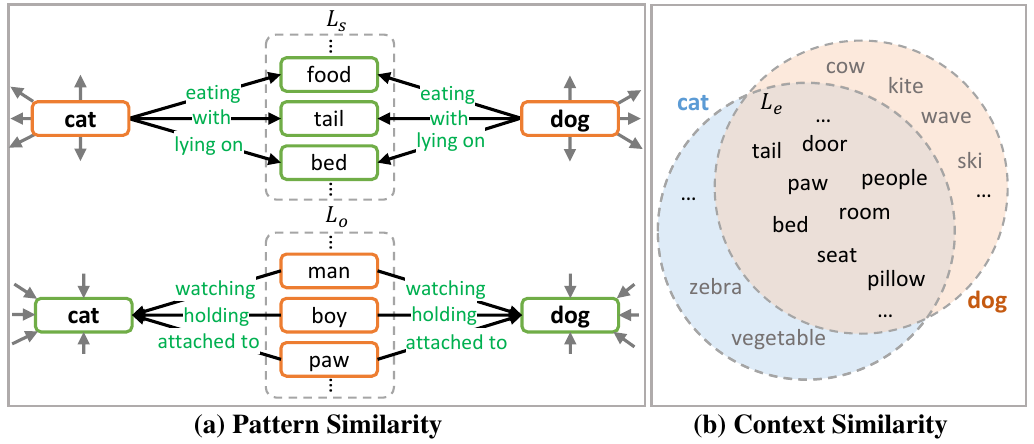}
   
  \caption{Illustration of the pattern and context similarity between entity categories \texttt{cat} and \texttt{dog}. a) White boxes are the behavior patterns common to two categories. b) Blue and red circles are entity categories that co-occur with \texttt{cat} and \texttt{dog}, respectively. 
  }
  \label{fig:cluster}
\end{figure}

\emph{\textrm{(i)} Pattern Similarity.}
It measures the overlap of the behavior patterns of two entity categories~\cite{wei2020hose}. Figure~\ref{fig:cluster}(a) visualizes the common pattern between \texttt{cat} and \texttt{dog} (\eg, they all can be attached by the \texttt{paw}). Pattern similarity between two entity categories $c_i$ and $c_j$ is defined as:
\begin{equation}
\begin{small}
    \begin{aligned}
        Si{m_p}({c_i},{c_j}) & = \frac{{|{L_s}|}}{{{d_{out}}({c_i}) + {d_{out}}({c_j}) - |{L_s}|}} \\
        & + 
        \frac{{|{L_o}|}}{{{d_{in}}({c_i}) 
        +  {d_{in}}({c_j}) - |{L_o}|}},
    \end{aligned}    
\end{small}
\end{equation}
where $|L_s|$ ($|L_o|$) is the number of common \texttt{pred}-\texttt{obj} (\texttt{sub}-\texttt{pred}) classes of the triplet whose \texttt{sub} (\texttt{obj}) class is $c_i$ or $c_j$. $d_{in}(c_i)$ ($d_{out}(c_i)$) is the number of incoming (outgoing) edges of the entity with class $c_i$ in the whole dataset.

\emph{\textrm{(ii)} Context Similarity.}
It measures the overlap of other entity categories in images of the two entity categories. Both \texttt{cat} and \texttt{dog} can appear in the same scene with entity categories, \eg, \texttt{room} in Figure~\ref{fig:cluster}(b). It is defined as: 
\begin{equation}
\begin{small}
    Si{m_c}({c_i},{c_j}) = \frac{{|{L_e}|}}{{{d_{co}}({c_i}) + {d_{co}}({c_j}) - |{L_e}|}}, \\
\end{small}
\end{equation}
where $|L_e|$ denotes the number of entity category intersections that appear in the same image with entity category $c_i$ or with $c_j$. $d_{co}(c_i)$ is the number of entity instances in the whole dataset that appear in the same image as category $c_i$.

\emph{\textrm{(iii)} Semantic Similarity.}
It is the cosine similarity between two entity categories on semantic embedding~\cite{pennington2014glove} space:
\begin{equation}
    Si{m_s}({c_i},{c_j})  = \frac{{{\bm{w}_i}\bigcdot{\bm{w}_j}}}{{|{\bm{w}_i}||\bm{w}j|}},
\end{equation}
The final similarity function $Sim({c_i},{c_j})$ used in clustering is the weighted sum of above three similarities. Complete clustering algorithm and results are in the \textbf{appendix}.

\noindent\textbf{Entity Feature Selection and Replacement.} This step selects an entity feature from the feature bank based on the selected entity category and replaces it with the query triplet. Specifically, we first update a new entity category of query triplet (\eg, \texttt{cat}-\texttt{laying on}-\texttt{bed}). Then, we select all triplet features with the same category as query triplet from feature bank as candidates (\cf. Figure~\ref{fig:cfa}(a)). And we utilize spatial restriction to filter unreasonable triplet features. 

\emph{Spatial Restriction.} In real augmentation, even if the two entity categories appear to be interchangeable, the positions of the subject and object after replacement may not make sense. For example, in VG dataset (\cf. \textbf{appendix}), though category \texttt{hat} and \texttt{shoe} are in the same cluster, directly replacing \texttt{hat} with \texttt{shoe} would place the \texttt{shoe} on the \texttt{head} and make it contrary to common sense.

To further ensure the rationality of the replacement of the two entity features, we use the cosine similarity of the relative spatial position of subject-object of the query triplet and the selected triplet from the feature bank as the restriction:
\begin{equation}
    {{Si}}{{{m}}_d}({\bm{p}_i},{\bm{p}_j}) = \frac{{{\bm{p}_i}\bigcdot{\bm{p}_j}}}{{|{\bm{p}_i}||\bm{p}j|}},
\end{equation}
where $\bm{p}_{i}$ is the spatial vector from the center of the subject bbox to the center of object bbox. When the ${Sim}_d$ between $\bm{p}_i$ of the query triplet and $\bm{p}_j$ of the selected triplet from the feature bank is larger than the threshold $\sigma$, the entity feature of the selected triplet is reasonable for replacement.

During replacement, we randomly select one of the reasonable entity features and replace it into query triplet and change the original entity target to the new entity category.

\subsubsection{Extrinsic-CFA}
\label{sec:exc}

Extrinsic-CFA aims to enrich the feature of the query triplet through the extrinsic information (\ie, the context formed by all the entities of an image) of context triplets. Different from the Intrinsic-CFA, we first select context triplet during training for computation efficiency. Extrinsic-CFA includes three steps: context triplet selection, query triplet feature selection, and feature augmentation (\cf.~Figure~\ref{fig:cfa}(b)). Then, we will detailedly introduce each step of Extrinsic-CFA. 

\noindent\textbf{Context Triplet Selection.} For each image, this step selects context triplets for extrinsic-CFA. Specifically, we randomly select context triplets with foreground or background predicate categories to provide more extrinsic information. \emph{For foreground context triplets}, considering the large variation in the number of triplet samples for different predicate categories, we randomly sample the context triplets by using probability $p = (\eta - {\eta_{r}})/\eta \times \gamma$ during training, where $\gamma$ is hyperparameter. \emph{For background context triplets}, we randomly select them with the same subject-object pair as the tail predicate triplet in the whole dataset.

\noindent\textbf{Query Triplet Feature Selection.} This step selects a query triplet with its features for the feature augmentation. Specifically, we select triplets with the same subject-object categories (\eg, \texttt{pillow}-\texttt{bed}) as the context triplet (\eg, \texttt{pillow}-\texttt{on}-\texttt{bed}) from feature bank as the candidate query triplets. Then, we implement the same spatial restriction as intrinsic-CFA to filter unreasonable triplets. Finally, we randomly select one of the reasonable triplets as the query triplet (\eg, \texttt{pillow}-\texttt{laying on}-\texttt{bed}), and its features are utilized for augmentation (\cf.~Figure~\ref{fig:cfa}(b)).

\noindent\textbf{Feature Augmentation.} To reduce the impact on other triplets of the original image and enhance the extrinsic features of the tail predicate triplets, we perform mixup operation~\cite{zhang2017mixup} between the selected query triplet (\eg, \texttt{pillow}-\texttt{laying on}-\texttt{bed}) and context triplet (\eg, \texttt{pillow}-\texttt{on}-\texttt{bed}), \cf.~Figure~\ref{fig:cfa}(b). The mixup operation is written as:
\begin{gather}
    \tilde{\bm{v}}_s = \theta \bm{v}_s + (1 - \theta ){{\bm{v}}_s^{\prime}}, \\
    \tilde{\bm{v}}_o = \theta {\bm{v}}_o + (1 - \theta ){{\bm{v}}_o^{\prime}}, \\
    \tilde{\bm{u}} = \theta \bm{u} + (1 - \theta ){{\bm{u}}^{\prime}},
\end{gather}
where $\bm{v}_s^{\prime}$ ($\bm{v}_s$) and $\bm{v}_o^{\prime}$ ($\bm{v}_o$) denote the visual features of the subject and object of the query triplet (context triplet), respectively. The $\bm{u}^{\prime}$ ($\bm{u}$) denotes the visual feature of the uion box of the query triplet (context triplet). Similarly, predicate target ground-truth $r$ of the context triplet is mixed:
\begin{equation}
        \tilde{r} = \theta {r} + (1 - \theta ){r^{\prime}},
\end{equation}
where $\theta$ is utilized to control the degree of mixup operation.

After the mixup operation, the query triplet feature can be enhanced in context modeling (\cf.~Figure~\ref{fig:framework}) by leveraging the extrinsic features of the context triplet.

\subsection{Training Objectives}
\label{sec:train}

\noindent\textbf{Cross-Entropy Losses.} The optimization objective of commonly used SGG models mainly includes the cross-entropy of entity and relation classification, the loss functions are:
\begin{equation}
 {L_{obj}} = \sum\nolimits_i {\mathtt{XE}({\widehat{o}_i},{o_i})}, \; {L_{rel}} = \sum\nolimits_{ij} {\mathtt{XE}({\widehat{r}_{ij}},{r_{ij}})},
\end{equation}
where ${\widehat{o}_i}$ is the predicted entity category and ${o}_i$ is the ground-truth entity category. ${\widehat{r}_{ij}}$ is the predicted predicate category and ${r}_{ij}$ is the ground-truth predicate category.

\noindent\textbf{Contrastive Loss.} Due to the pattern changes between the entity features after mixup and original entity features, using only XE losses may result in performance drops in entity classification. To maintain the discriminative entity features after mixup, we further apply a contrastive loss~\cite{chen2020simple}:
\begin{equation}
L_{cl} =  - \log \frac{{\exp (Si{m_e}({{\bm{z}}_i},{\bm{z}}_j)/\tau )}}{{\sum\nolimits_{k = 1}^{2M} {{1_{[k \ne i]}}\exp (Si{m_e}({{\bm{z}}_i},{{\bm{z}}_k})/\tau )} }},
\end{equation}
where ${\bm{z}}_i$ and ${\bm{z}}_j$ represent the output of the layer before predictor of original entity and the entity after mixup operation, $M$ is the number of entities in which the mixup operation is performed. $Si{m_e(\bigcdot,\bigcdot)}$ is the cosine similarity.

\noindent\textbf{Training.} During training, the total loss includes the cross-entropy losses $L_{obj}$, $L_{rel}$ and the extra contrastive loss $L_{cl}$:
\begin{equation}
L_{total} = L_{rel} + L_{obj} + \beta L_{cl},
\end{equation}
where $\beta$ is used to regulate the magnitude of loss.

\addtolength{\tabcolsep}{-1.5pt}
\begin{table*}[!ht]
  \centering
    \scalebox{0.8}{
    \begin{tabular}{l|cc|cc|c|cc|cc|c|cc|cc|c}
    \hline
    \multicolumn{1}{c|}{\multirow{3}[4]{*}{SGG Models}} & \multicolumn{5}{c|}{\textbf{PredCls}}   & \multicolumn{5}{c|}{\textbf{SGCls}}    & \multicolumn{5}{c}{\textbf{SGGen}} \\
    \cline{2-16}    \multicolumn{1}{c|}{} & \multicolumn{2}{c|}{mR@K} & \multicolumn{2}{c|}{\textcolor{gray}{R@K}} & \multicolumn{1}{c|}{\multirow{2}[2]{*}{Mean}} & \multicolumn{2}{c|}{mR@K} & \multicolumn{2}{c|}{\textcolor{gray}{R@K}} & \multicolumn{1}{c|}{\multirow{2}[2]{*}{Mean}} & \multicolumn{2}{c|}{mR@K} & \multicolumn{2}{c|}{\textcolor{gray}{R@K}} & \multicolumn{1}{c}{\multirow{2}[2]{*}{Mean}} \\
    \multicolumn{1}{c|}{} & \multicolumn{1}{c}{50} & \multicolumn{1}{c|}{100} & \multicolumn{1}{c}{\textcolor{gray}{50}} & \multicolumn{1}{c|}{\textcolor{gray}{100}} &       & \multicolumn{1}{c}{50} & \multicolumn{1}{c|}{100} & \multicolumn{1}{c}{\textcolor{gray}{50}} & \multicolumn{1}{c|}{\textcolor{gray}{100}} &       & \multicolumn{1}{c}{50} & \multicolumn{1}{c|}{100} & \multicolumn{1}{c}{\textcolor{gray}{50}} & \multicolumn{1}{c|}{\textcolor{gray}{100}} &  \\
    \hline
        \textcolor{gray}{Motifs~~\cite{zellers2018neural}$_{\textit{CVPR'18}}$}
        & \textcolor{gray}{16.5} & \textcolor{gray}{17.8}& \textcolor{gray}{65.5} & \textcolor{gray}{67.2} & \textcolor{gray}{41.8}
        & \textcolor{gray}{8.7}  & \textcolor{gray}{9.3}  & \textcolor{gray}{39.0} & \textcolor{gray}{39.7} & \textcolor{gray}{24.2}
        & \textcolor{gray}{5.5}  & \textcolor{gray}{6.8}  & \textcolor{gray}{32.1} & \textcolor{gray}{36.9} & \textcolor{gray}{20.3} \\ 
        \textcolor{gray}{VCTree~\cite{tang2019learning}$_{\textit{CVPR'19}}$}
        & \textcolor{gray}{17.1} & \textcolor{gray}{18.4} & \textcolor{gray}{65.9} & \textcolor{gray}{67.5} & \textcolor{gray}{42.2}
        & \textcolor{gray}{10.8} & \textcolor{gray}{11.5} & \textcolor{gray}{45.6} & \textcolor{gray}{46.5} & \textcolor{gray}{28.6} 
        & \textcolor{gray}{7.2}  & \textcolor{gray}{8.4}  & \textcolor{gray}{32.0} & \textcolor{gray}{36.2} & \textcolor{gray}{20.9} \\
        \textcolor{gray}{Transformer~\cite{vaswani2017attention}$_{\textit{NIPS'17}}$}
        & \textcolor{gray}{17.9} & \textcolor{gray}{19.6} & \textcolor{gray}{63.6} & \textcolor{gray}{65.7} & \textcolor{gray}{41.7}
        & \textcolor{gray}{9.9}  & \textcolor{gray}{10.5} & \textcolor{gray}{38.1} & \textcolor{gray}{39.2} & \textcolor{gray}{24.4}
        & \textcolor{gray}{7.4}  & \textcolor{gray}{8.8}  & \textcolor{gray}{30.0} & \textcolor{gray}{34.3} & \textcolor{gray}{20.1} \\
        \textcolor{gray}{BGNN~\cite{li2021bipartite}$_{\textit{CVPR'21}}$}
        & \textcolor{gray}{30.4} & \textcolor{gray}{32.9} & \textcolor{gray}{59.2} & \textcolor{gray}{61.3} & \textcolor{gray}{45.9} 
        & \textcolor{gray}{14.3} & \textcolor{gray}{16.5} & \textcolor{gray}{37.4} & \textcolor{gray}{38.5} & \textcolor{gray}{26.7} 
        & \textcolor{gray}{10.7} & \textcolor{gray}{12.6} & \textcolor{gray}{31.0} & \textcolor{gray}{35.8} & \textcolor{gray}{22.5} \\
        \hline
        Motifs+PCPL~\cite{yan2020pcpl}$_{\textit{ACMMM'20}}$
        & 24.3 & 26.1 & \textcolor{gray}{54.7} & \textcolor{gray}{56.5} & 40.4
        & 12.0 & 12.7 & \textcolor{gray}{35.3} & \textcolor{gray}{36.1} & 24.0 
        & 10.7 & 12.6 & \textcolor{gray}{27.8} & \textcolor{gray}{31.7} & 20.7 \\  
    Motifs+DLFE~\cite{chiou2021recovering}$_{\textit{ACMMM'21}}$
        & 26.9 & 28.8 & \textcolor{gray}{52.5} & \textcolor{gray}{54.2} & 40.6
        & 15.2 & 15.9 & \textcolor{gray}{32.3} & \textcolor{gray}{33.1} & 24.1
        & 11.7 & 13.8 & \textcolor{gray}{25.4} & \textcolor{gray}{29.4} & 20.1 \\
        Motifs+BPL-SA~\cite{guo2021general}$_{\textit{ICCV'21}}$
        &  29.7 & 31.7 & \textcolor{gray}{50.7} & \textcolor{gray}{52.5} & 41.2
        &  16.5 & 17.5 & \textcolor{gray}{30.1} & \textcolor{gray}{31.0} & 23.8
        &  13.5 & 15.6 & \textcolor{gray}{23.0} & \textcolor{gray}{26.9} & 19.8 \\ 
        Motifs+NICE~\cite{li2022devil}$_{\textit{CVPR'22}}$
        &  29.9 & 32.3 & \textcolor{gray}{55.1} & \textcolor{gray}{57.2} & 43.6
        &  16.6 & 17.9 & \textcolor{gray}{33.1} & \textcolor{gray}{34.0} & 25.4
        &  12.2 & 14.4 & \textcolor{gray}{27.8} & \textcolor{gray}{31.8} & 21.6 \\ 
        Motifs+IETrans~\cite{zhang2022fine}$_{\textit{ECCV'22}}$
        &  30.9 & 33.6 & \textcolor{gray}{54.7} & \textcolor{gray}{56.7} & 44.0
        &  16.8 & 17.9 & \textcolor{gray}{32.5} & \textcolor{gray}{33.4} & 25.2
        &  12.4 & 14.9 & \textcolor{gray}{26.4} & \textcolor{gray}{30.6} & 21.1 \\ 
         {\textbf{Motifs+CFA (ours)}}
        &  35.7 &  38.2 &   \textcolor{gray}{54.1} &  \textcolor{gray}{56.6} &  \textbf{46.2}&  17.0 &  18.4&  \textcolor{gray}{34.9} &  \textcolor{gray}{36.1} & 
         {\textbf{26.6}} &  13.2 & 15.5 &   \textcolor{gray}{27.4} &  \textcolor{gray}{31.8} &  \textbf{22.0} \\
        \hline

        VCTree+PCPL~\cite{yan2020pcpl}$_{\textit{ACMMM'20}}$
        & 22.8 & 24.5 &  \textcolor{gray}{56.9} & \textcolor{gray}{58.7} & 40.7
        & 15.2 & 16.1 &  \textcolor{gray}{40.6} & \textcolor{gray}{41.7} & 28.4 
        & 10.8 & 12.6 &  \textcolor{gray}{26.6} & \textcolor{gray}{30.3} & 20.1 \\
    VCTree+DLFE~\cite{chiou2021recovering}$_{\textit{ACMMM'21}}$
        & 25.3 & 27.1 & \textcolor{gray}{51.8} & \textcolor{gray}{53.5} & 39.4 
        & 18.9 & 20.0 & \textcolor{gray}{33.5} & \textcolor{gray}{34.6} & 26.8 
        & 11.8 & 13.8 & \textcolor{gray}{22.7} & \textcolor{gray}{26.3} & 18.7 \\
        VCTree+BPL-SA~\cite{guo2021general}$_{\textit{ICCV'21}}$
        &  30.6 & 32.6 & \textcolor{gray}{50.0} & \textcolor{gray}{51.8} & 41.3
        &  20.1 & 21.2 & \textcolor{gray}{34.0} & \textcolor{gray}{35.0} & 27.6
        &  13.5 & 15.7 & \textcolor{gray}{21.7} & \textcolor{gray}{25.5} & 19.1 \\ 
        VCTree+NICE~\cite{li2022devil}$_{\textit{CVPR'22}}$
        &  30.7 & 33.0 & \textcolor{gray}{55.0} & \textcolor{gray}{56.9} & 43.9
        &  19.9 & 21.3 & \textcolor{gray}{37.8} & \textcolor{gray}{39.0} & 29.5
        &  11.9 & 14.1 & \textcolor{gray}{27.0} & \textcolor{gray}{30.8} & 21.0 \\

        VCTree+IETrans~\cite{zhang2022fine}$_{\textit{ECCV'22}}$
        &  30.3 & 33.9 & \textcolor{gray}{53.0} & \textcolor{gray}{55.0} & 43.1
        &  16.5 & 18.1 & \textcolor{gray}{32.9} & \textcolor{gray}{33.8} & 25.3
        &  11.5 & 14.0 & \textcolor{gray}{25.4} & \textcolor{gray}{29.3} & 20.1 \\ 
        \textbf{VCTree+CFA (ours)}
        & 34.5 & 37.2 &  \textcolor{gray}{54.7} &  \textcolor{gray}{57.5} & \textbf{46.0} &  19.1 &  20.8&  \textcolor{gray}{42.4} &  \textcolor{gray}{43.5} & 
         \textbf{31.5} &  13.1 &  15.5 &   \textcolor{gray}{27.1} &  \textcolor{gray}{31.2} &  \textbf{21.7} \\
        \hline
    Transformer+IETrans~\cite{zhang2022fine}$_{\textit{ECCV'22}}$
        & 30.8 & 34.5 & \textcolor{gray}{51.8} & \textcolor{gray}{53.8} & 42.7 
        & 17.4 & 19.1 & \textcolor{gray}{32.6} & \textcolor{gray}{33.5} & 25.7 
        & 12.5 & 15.0 & \textcolor{gray}{25.5} & \textcolor{gray}{29.6} & 20.7 \\
         {\textbf{Transformer+CFA (ours)}}
        &  30.1 & 33.7 &   \textcolor{gray}{59.2} &  \textcolor{gray}{61.5} &  \textbf{46.1} &  {15.7} &  {17.2}&  \textcolor{gray}{36.3} &  \textcolor{gray}{37.3} & 
         \textbf{26.6} &  {12.3} &  {14.6} &  \textcolor{gray}{27.7} &  \textcolor{gray}{32.1} &  \textbf{21.7} \\
        \hline
    \end{tabular}
    }
    \vspace{-0.5em}
    \caption{Performance (\%) of the SOTA trade-off SGG models on VG~\cite{krishna2017visual}. ``Mean" is the average of mR@50/100 and R@50/100.}
      
    \label{tab:compare_with_sota}
\end{table*}%
\addtolength{\tabcolsep}{1.5pt}

\addtolength{\tabcolsep}{-1.5pt}
\begin{table*}[!ht]
  \centering
    \scalebox{0.8}{
    \begin{tabular}{l|ccc|ccc|ccc}
    \hline
    \multicolumn{1}{c|}{\multirow{2}[2]{*}{SGG Models}} & \multicolumn{3}{c|}{\textbf{PredCls}} & \multicolumn{3}{c|}{\textbf{SGCls}} & \multicolumn{3}{c}{\textbf{SGGen}} \\
    \cline{2-10} \multicolumn{1}{c|}{} & \multicolumn{1}{c|}{mR@20} & \multicolumn{1}{c|}{mR@50} & \multicolumn{1}{c|}{mR@100} & \multicolumn{1}{c|}{mR@20} & \multicolumn{1}{c|}{mR@50} & \multicolumn{1}{c|}{mR@100} & \multicolumn{1}{c|}{mR@20} & \multicolumn{1}{c|}{mR@50} & \multicolumn{1}{c}{mR@100} \\
    \hline
        Motifs+TDE~\cite{tang2020unbiased}$_{\textit{CVPR'20}}$
        & 18.5 & 25.5 & 29.1
        & 9.8 & 13.1 & 14.9
        & 5.8 & 8.2 & 9.8 \\
        Motifs+CogTree~\cite{yu2021cogtree}$_{\textit{IJCAI'21}}$
        & 20.9 & 26.4 & 29.0 
        & 12.1 & 14.9 & 16.1 
        & 7.9 & 10.4 & 11.8  \\ 
        Motifs+RTPB~\cite{chen2022resistance}$_{\textit{AAAI'22}}$
        & 28.8 & 35.3 & 37.7 
        & 16.3 & 20.0 & 21.0  
        & 9.7 & 13.1 & 15.5  \\
        Motifs+PPDL~\cite{li2022ppdl}$_{\textit{CVPR'22}}$
        & 27.9 & 32.2 & 33.3 
        & 15.8 & 17.5 & 18.2 
        & 9.2 & 11.4 & 13.5  \\ 
        Motifs+GCL~\cite{dong2022stacked}$_{\textit{CVPR'22}}$
        & 30.5 & 36.1 & 38.2  
        & \textbf{18.0} & 20.8 & 21.8  
        & \textbf{12.9} & \textbf{16.8} & \textbf{19.3} \\ 
        Motif+HML~\cite{deng2022hierarchical}$_{\textit{ECCV'22}}$
        & 30.1 & 36.3 & 38.7 
        & 17.1 & 20.8 & 22.1 
        & 10.8 & 14.6 & 17.3  \\
         {\textbf{Motifs+CFA$^{\ddag}$ (ours)}}
        &   \textbf{31.5} &    \textbf{39.9} &  \textbf{43.0} 
        &   {17.3} &   \textbf{20.9} &  \textbf{22.4}
        &   {11.2} &  {15.3} &  {18.1} \\
        \hline
        VCTree+TDE~\cite{tang2020unbiased}$_{\textit{CVPR'20}}$
        & 18.4 & 25.4 & 28.7
        & 8.9 & 12.2 & 14.0
        & 6.9 & 9.3 & 11.1 \\
        VCTree+CogTree~\cite{yu2021cogtree}$_{\textit{IJCAI'21}}$
        & 22.0 & 27.6 & 29.7 
        & 15.4 & 18.8 & 19.9 
        & 7.8 & 10.4 & 12.1  \\ 
        VCTree+RTPB~\cite{chen2022resistance}$_{\textit{AAAI'22}}$
        & 27.3 & 33.4 & 35.6 
        & 20.6 & 24.5 & 25.8 
        & 9.6 & 12.8 & 15.1  \\ 
        VCTree+PPDL~\cite{li2022ppdl}$_{\textit{CVPR'22}}$
        & 29.7 & 33.3 & 33.8 
        & 20.3 & 21.8 & 22.4 
        & 9.1 & 11.3 & 13.3  \\ 
        VCTree+GCL~\cite{dong2022stacked}$_{\textit{CVPR'22}}$
        & 31.4 & 37.1 & 39.1 
        & 19.5 & 22.5 & 23.5 
        & \textbf{11.9} & \textbf{15.2} & 17.5  \\
        VCTree+HML~\cite{deng2022hierarchical}$_{\textit{ECCV'22}}$
        & 31.0 & 36.9  & 39.2 
        & 20.5 & 25.0  & 26.8 	
        & 10.1 & 13.7  & 16.3 \\
         {\textbf{VCTree+CFA$^{\ddag}$ (ours)}} 
        &   \textbf{31.6} &  \textbf{39.2} &  \textbf{42.5}   
        &   \textbf{21.5} &  \textbf{26.3} &  \textbf{28.3} 
        &   {10.8} &  {15.1} &  \textbf{17.9} \\
        \hline
    Transformer+CogTree~\cite{yu2021cogtree}$_{\textit{IJCAI'21}}$ 
        & 22.9 & 28.4 & 31.0  
        & 13.0 & 15.7 & 16.7 
        & 7.9 & 11.1 & 12.7  \\
Transformer+HML~\cite{deng2022hierarchical}$_{\textit{ECCV'22}}$ 
        & 27.4 & 33.3 & 35.9  
        & 15.7 & 19.1 & 20.4 
        & \textbf{11.4} & 15.0 & 17.7  \\
       {\textbf{Transformer+CFA$^{\ddag}$ (ours)}} 
        &  \textbf{31.2} &  \textbf{38.6} &  \textbf{41.5} 
        &  \textbf{17.2} &  \textbf{20.9} &  \textbf{22.7}
        &  {10.6} &  \textbf{15.0} &  \textbf{17.9} \\
        \hline
    \end{tabular}
}
          \vspace{-0.5em}
    \caption{Performance (\%) of the SOTA tail-focused SGG models on VG~\cite{krishna2017visual}. $\ddag$ means using the component prior knowledge.}
     
    \label{tab:compare_with_sota_tail}
\end{table*}%
\addtolength{\tabcolsep}{1.5pt}

\addtolength{\tabcolsep}{-1.5pt}
\begin{table}[htbp]
  \centering
      \scalebox{0.8}{
    \begin{tabular}{|l|c|c|c|}
       \hline
    \multicolumn{1}{|c|}{\multirow{2}[2]{*}{Models}} & {PredCls} & {SGCls} & {SGGen} \\
\cline{2-4} & \multicolumn{1}{c|}{mR@50/100}  & \multicolumn{1}{c|}{mR@50/100}  & \multicolumn{1}{c|}{mR@50/100} \\
    \hline
    Motifs~\cite{zellers2018neural} & 13.9  / 14.7  & 7.2 / 7.5   & 5.5 / 6.6 \\
   ~+\textbf{CFA} & \textbf{31.7} / \textbf{33.8}  & \textbf{14.2} / \textbf{15.2}  & \textbf{11.6} / \textbf{13.2} \\
   \hline
    VCTree~\cite{tang2019learning} & 14.4 / 15.3  & 6.1 / 6.6   & 5.8  / 6.0 \\
    ~+\textbf{CFA} & \textbf{33.4}  / \textbf{35.1}  & \textbf{14.1} / \textbf{15.0}  & \textbf{10.8} / \textbf{12.6} \\
    \hline
    Transformer~\cite{vaswani2017attention} & 15.2  /16.1  & 7.5  / 7.9   & 6.9  / 7.8 \\
   ~+\textbf{CFA} & \textbf{27.8}  / \textbf{29.4}  & \textbf{16.2} / \textbf{16.9}  & \textbf{13.4} / \textbf{15.3} \\
   \hline
    \end{tabular}%
    }
      \caption{Performance (\%) of the SGG models on GQA~\cite{hudson2019gqa}.}
    
  \label{tab:gqa}%
\end{table}%
\addtolength{\tabcolsep}{1.5pt}

\section{Experiments}
\subsection{Experimental Settings and Details}

\begin{figure*}[htbp]
    \begin{minipage}[l]{0.6\linewidth}
        \centering
        \includegraphics[width=\linewidth]{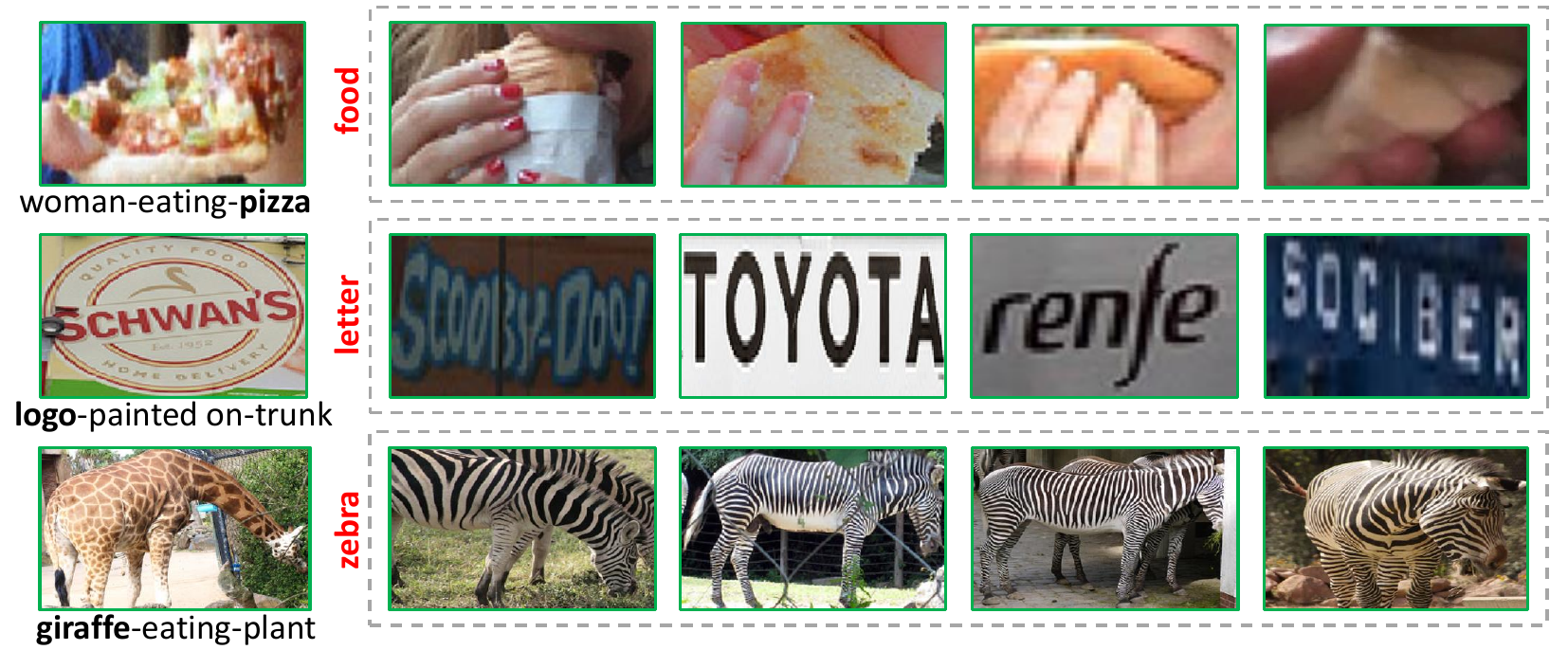}
         
        \caption{The examples of reasonable entities for the query triplet in intrinsic CFA. The lefts are original entities of the query triplets. The rights are some alternate entity categories (\textcolor{red}{red}) and their samples (\textcolor{gray}{gray} boxes) for each query triplet to replace.}
        \label{fig:example}
    \end{minipage}\hspace{0.5em}
    \begin{minipage}[!r]{0.4\textwidth}
      \captionsetup{type=table}
        \centering
       
    \scalebox{0.85}{
    \setlength\tabcolsep{5pt} 
    \begin{tabular}{|ccc|c|c|c|}
    \specialrule{0em}{0.5pt}{0.5pt}
    \hline
    \multicolumn{3}{|c|}{Component} & \multicolumn{3}{c|}{PredCls} \\
    \hline
    \footnotesize{IN} & \footnotesize{EX-fg} & \footnotesize{EX-bg} & \footnotesize{mR@50 / 100} & \footnotesize{\textcolor{gray}{R@50 / 100}} & \footnotesize{Mean} \\
    \hline
               &            &            &   16.5 / 17.8 &  \textcolor{gray}{65.6 / 67.2}  & 41.8 \\
    \checkmark &            &            & 19.3	/ 21.2	
               & \textcolor{gray}{64.7 / 66.6} & 43.0 \\
               & \checkmark &            & 25.6	/ 27.8 
               & \textcolor{gray}{63.0 / 64.8} & 45.3 \\
               &            & \checkmark & 23.9	/ 26.3 
               & \textcolor{gray}{63.3 / 65.7} & 44.8 \\
    \checkmark & \checkmark &            & 27.2	/ 29.3
               & \textcolor{gray}{61.9 / 64.3} & 45.7 \\
    \checkmark &            & \checkmark & 27.5 / 30.0	
               & \textcolor{gray}{61.7 / 64.1} & 45.8 \\
               & \checkmark & \checkmark & 27.8 / 30.3 	    
               & \textcolor{gray}{60.7 / 63.4} & 45.6 \\
    \checkmark & \checkmark & \checkmark & \textbf{35.7}	/ \textbf{38.2} & \textcolor{gray}{54.1 / 56.6} & \textbf{46.2} \\
    \hline
    \end{tabular}%
    }
      
    \caption{Ablation study on each component on VG~\cite{krishna2017visual}. IN: Replace the intrinsic features. Ex-fg: Mix up the extrinsic features of foreground triplets. Ex-bg: Mix up the extrinsic features of background triplets.}
     
    \label{tab:component}%
  \end{minipage}
\end{figure*}

\noindent\textbf{Tasks.} We evaluated models in three tasks~\cite{xu2017scene}: 1) \emph{Predicate Classification} (\textbf{PredCls}): Predicting the predicate category given all ground-truth entity bboxes and categories. 2) \emph{Scene Graph Classification} (\textbf{SGCls}): Predicting categories of the predicate and entity given all ground-truth entity bboxes. 3) \emph{Scene Graph Generation} (\textbf{SGGen}): Detecting all entities and their pairwise predicates.

\noindent\textbf{Metrics.} We evaluated SGG models on three metrics: 1) \emph{Recall@K} (\textbf{R@K}): It indicates the proportion of ground-truths that appear among the top-$K$ confident predicted relation triplets. 2) \emph{mean Recall@K} (\textbf{mR@K}): It is the average of R@K scores which are calculated for each predicate category separately. 3) \textbf{Mean}: It is the average of all R@K and mR@K scores. Since R@K favors head predicates while mR@K favors tail predicates, the Mean can better reflect the overall performance of all predicates~\cite{li2022devil}.

\noindent\textbf{Datasets and Implementation Details.} We conducted all experiments on two datasets: \textbf{VG}~\cite{krishna2017visual} and \textbf{GQA}~\cite{hudson2019gqa}. More details of datasets and implementation are in the appendix. 

\subsection{Comparison with State-of-the-Arts}

\textbf{Setting.} Due to the model-agnostic nature, we equipped our CFA with three strong two-stage SGG baselines: Motifs~\cite{zellers2018neural}, VCTree~\cite{tang2019learning} and Transformer~\cite{vaswani2017attention}, and they are denoted as \textbf{Motifs+CFA}, \textbf{VCTree+CFA}, and \textbf{Transformer+CFA}, respectively. In addition, due to the limited diversity of tail predicate components~\cite{li2022rethinking}, it has a high correlation with the category of subject \& object. Thus, we further equip our three models with the component prior knowledge collected from the dataset to further improve the performance of the tail predicates. And they are denoted as \textbf{Motifs+CFA$^\ddag$}, \textbf{VCTree+CFA$^\ddag$}, and \textbf{Transformer+CFA$^\ddag$} (More details about the priors are discussed in appendix). 

\textbf{Baselines.} We compared our methods with the SOTA models in the VG dataset (Table~\ref{tab:compare_with_sota}, Table~\ref{tab:compare_with_sota_tail}) and the GQA dataset (Table~\ref{tab:gqa}). Specifically, these models can be divided into three groups: 1) \textbf{Model-specific designs}: Motifs, VCTree, Transformer, and BGNN~\cite{li2021bipartite}. 2) \textbf{Model-agnostic trade-off} methods: consider the performance of all predicates comprehensively (\ie, higher Mean), \eg, PCPL~\cite{yan2020pcpl}, DLFE~\cite{chiou2021recovering}, BP-LSA~\cite{guo2021general}, NICE~\cite{li2022devil}, and IETrans~\cite{zhang2022fine}. 3) \textbf{Model-agnostic tail-focused} methods: improve tail predicates performance at the expense of excessively sacrificing the head (\ie, higher mR@K and R@50 less than 50.0\% on PredCls), \eg, TDE~\cite{tang2020unbiased}, CogTree~\cite{yu2021cogtree}, RTPB~\cite{chen2022resistance}, PPDL~\cite{li2022ppdl}, GCL~\cite{dong2022stacked}, and HML~\cite{deng2022hierarchical}. For a fair comparison, we compared with the methods in the last two groups.

\textbf{Quantitative Results on VG.} 
From the results of \textbf{trade}\textbf{-off} methods in Table~\ref{tab:compare_with_sota}, we can observe that: 1) Compared to the three strong baselines (\ie, Motifs, VCTree and Transformer), CFA can significantly improve model performance on mR@K metric over all three settings. 2) CFA can achieve the best trade-off between R@K and mR@K, \ie, highest Mean, and surpass the SOTA trade-off method NICE~\cite{li2022devil} in Mean metric under all settings. CFA shows minimal performance degradation on the head predicates (\cf.~R@K) while maintaining the performance of the tail (\cf.~mR@K), demonstrating the superiority of CFA considering all predicates. From the results of \textbf{tail-focused} methods in Table~\ref{tab:compare_with_sota_tail}, we can observe that: after further implementing strategies to improve tail performance, CFA$^\ddag$ can achieve the highest mR@K and exceed the SOTA tail-focused method HML~\cite{deng2022hierarchical} on mR@K metric under all settings. More experiment analysis is in the appendix.

\textbf{Quantitative Results on GQA.} From the results of Table~\ref{tab:gqa}, we can observe that: CFA can also greatly improve the mR@K of three strong baselines (\ie, Motif, VCTree and Transformer) on the large dataset GQA, which proves the universality and effectiveness of our method.

\subsection{Ablation Studies}

\addtolength{\tabcolsep}{-1.5pt}
\begin{table}[!t]
        \centering
        \captionsetup{type=table} 
         
        \subfloat[Ablation study on each similarity in clustering of intrinsic CFA.]{
        \tablestyle{5pt}{1}    
        \scalebox{0.85}{
        \begin{tabular}{|ccc|c|c|c|}
    \hline
    \multicolumn{3}{|c|}{Similarity} & \multicolumn{3}{c|}{PredCls} \\
    \hline
    \footnotesize{Pattern} & \footnotesize{Context} & \footnotesize{Semantic} & \footnotesize{mR@50 / 100} & \footnotesize{\textcolor{gray}{R@50 / 100}} & \footnotesize{Mean} \\
    \hline
    \checkmark &            &            & 33.4 / 36.2 &              \textcolor{gray}{52.0 / 54.0}  & 43.9 \\
               & \checkmark &            & 34.6 / 37.3 & \textcolor{gray}{55.2 / 57.0}  & 46.0 \\
               &            & \checkmark & 35.3 / 38.0 &
               \textcolor{gray}{51.4 / 54.1}  & 44.7 \\
    \checkmark & \checkmark &            & 35.3 / 37.8 &              \textcolor{gray}{52.6 / 55.2}  & 45.2 \\
    \checkmark &            & \checkmark & 35.2 / 37.9 &              \textcolor{gray}{54.0 / 56.2}  & 45.8 \\
               & \checkmark & \checkmark & 34.5 / 36.9 &
               \textcolor{gray}{55.2 / 57.5}  & 46.0  \\
    \checkmark & \checkmark & \checkmark & \textbf{35.7} / \textbf{38.2} &              \textcolor{gray}{54.1 / 56.6}  & \textbf{46.2}  \\
    \hline
    \end{tabular}}}
    
    \subfloat[Ablation study on the number of clusters $K$ of intrinsic CFA.]{
        \tablestyle{4pt}{1.2}   
    \centering
    \scalebox{0.8}{
        \setlength\tabcolsep{2pt} \begin{tabular}{|c|c|c|c|}
    \hline
    \multirow{2}{*}{$K$} & \multicolumn{3}{c|}{PredCls} \\
    \cline{2-4}  
     & \footnotesize{mR@50 / 100}& \footnotesize{\textcolor{gray}{R@50 / 100}} & \footnotesize{Mean} \\
    \hline
    15  &   \textbf{35.7} / \textbf{38.2} &  \textcolor{gray}{54.1 / 56.6}  & \textbf{46.2} \\
    40  &   33.5 / 36.0 &  \textcolor{gray}{55.3 / 57.8}  & 45.7 \\
    150 &   33.2 / 35.7 &  \textcolor{gray}{55.7 / 57.9}  & 45.6 \\
    \hline
    \end{tabular}}}
    \subfloat[Ablation study on mixup parameter $\theta$ of extrinsic CFA.]{
		\tablestyle{4pt}{0.9}    \scalebox{0.8}{
		\setlength\tabcolsep{2pt}
		\begin{tabular}{|c|c|c|c|}
    \hline
    \multirow{2}{*}{$\theta$} & \multicolumn{3}{c|}{PredCls} \\
    \cline{2-4}  
     & \footnotesize{mR@50 / 100}& \footnotesize{\textcolor{gray}{R@50 / 100}} & \footnotesize{Mean} \\
    \hline
    0.0  & 27.1 / 29.5  & \textcolor{gray}{62.2 / 64.3}  & 45.8 \\ 
    0.3  & 28.5 / 31.0  & \textcolor{gray}{60.8 / 63.3}  & 45.9 \\
    0.5  & \textbf{35.7} / \textbf{38.2}  & \textcolor{gray}{54.1 / 56.6}  & \textbf{46.2} \\
    0.7  & 32.9 / 35.2  & \textcolor{gray}{50.4 / 53.0}  & 42.9 \\
    1.0  & 19.3 / 21.2  & \textcolor{gray}{64.7 / 66.6}  & 43.0 \\
    \hline
    \end{tabular}}}
      
    \caption{Ablation studies on the different hyperparameters of each component of CFA. Motifs~\cite{zellers2018neural} is used in all ablation on VG~\cite{krishna2017visual}.}
    	\label{tab:ablations}
      
\end{table}
\addtolength{\tabcolsep}{1.5pt}

\noindent\textbf{Effectiveness of Each Component.}
We evaluated the importance of each component of CFA based on Motifs~\cite{zellers2018neural} under the PredCls setting. There are three components of CFA: replace the intrinsic features (\textbf{IN}), mix up the extrinsic features of the foreground triplets (\textbf{EX-fg}), and mix up the extrinsic features of the background triplets (\textbf{EX-bg}). As reported in Table~\ref{tab:component}, we have the following observations: 1) Using only IN can slightly improve mR@K (\eg, 2.8\% $\sim$ 3.4\% gains) and slightly hurts R@K (0.6\% $\sim$ \% 0.9 loss). The reason is that there are not enough tail predicate triplets, resulting in the feature diversity still being limited after replacing the intrinsic features. 2) Both Ex-fg and Ex-bg can significantly improve mR@K and keep competitive R@K compared to baseline (\eg, 17.8\% \vs 30.3\% in mR@100, and 67.2\% \vs 63.4\% in R@100). 3) Combining all components allows for the best trade-off, \ie, the highest Mean.

\noindent\textbf{Similarity of Clustering in Intrinsic CFA.} We analyzed the influence of three kinds of similarity (pattern, context, and semantic similarities) under the PredCls setting with baseline model Motifs~\cite{zellers2018neural}. From Table~\ref{tab:ablations}(a), we can observe that using only pattern similarity shows the worst performance, since it ignores the contextual and semantic information which gives crucial guidance for selection. For example, both \texttt{boat} and \texttt{car} can be ridden, but we cannot replace \texttt{boat} with \texttt{car} because \texttt{car} can't appear in the sea. Once the other two similarities are fused, the SGG performance becomes robust and the highest Mean is achieved.

\noindent\textbf{Different Clusters $K$ of  Clustering in Intrinsic CFA.} We performed $K$ $\in$ \{15, 40, 150\} to evaluate the impact of the number of clusters under PredCls setting with Motifs~\cite{zellers2018neural}. All results are reported in Table~\ref{tab:ablations}(b). We find that $K$=15 works best in mR@K. The reason may be that when $K$=15, there is a larger selection range of replaceable entity categories and richer feature diversity of tail predicate triplets.

\noindent\textbf{Mixup Parameter $\theta$ in Extrinsic CFA.} As mentioned in Sec.~\ref{sec:exc}, $\theta$ indicates the proportion of selected features in final augmented features. We investigated $\theta$ $\in$ \{0.0, 0.3, 0.5, 0.7, 1.0\} under the PredCls setting with Motifs~\cite{zellers2018neural} in Table~\ref{tab:ablations}(c). When $\theta$ is too small, the query triplet has too much impact on the other triplets in the image of the context triplet, and when $\theta$ is too large, feature augmentation of the query triplet is not strong enough. To better trade-off the performance on different predicates, we set $\theta$ to $0.5$.

\noindent\textbf{Visualization of Reasonable Entities in Intrinsic CFA.} Figure~\ref{fig:example} shows some alternative entities for query triplets. The visual features of the candidate entities are quite different from the original entity, but are still reasonable for the query triplet, \eg, the samples of \texttt{zebra} differ greatly in color and texture from the \texttt{giraffe}, but they can all eat plants. Replacing the \texttt{giraffe} with the \texttt{zebra} can provide new features to enrich the feature diversity of \texttt{eating}.

\section{Conclusion and Future Work}
In this paper, we revealed the drawbacks of existing re-balancing methods and discovered that the key challenge for unbiased SGG is to learn proper decision boundaries under the severe long-tailed predicate distribution. Thus, we proposed a model-agnostic CFA framework that can enrich the feature space of tail categories by augmenting both intrinsic and extrinsic features of the relation triplets. Comprehensive experiments on the challenging VG and GQA datasets showed that CFA significantly improves the performance of unbiased SGG. In the future, we would like to extend CFA to compose new triplets features by fusing features from open-vocabulary categories or images from other domains.

\noindent\textbf{Acknowledgement.} This work was supported by the National Key Research \& Development Project of China (2021ZD0110700), the National Natural Science Foundation of China (U19B2043, 61976185),  and the Fundamental Research Funds for the Central Universities(226-2023-00048). Long Chen was supported by HKUST Special
Support for Young Faculty under Grant F0927.

{\small
\bibliographystyle{ieee_fullname}
\bibliography{egbib}
}

\end{document}